\documentclass[a4paper,conference]{IEEEtran}

\usepackage{amsmath,amssymb,amsfonts}
\usepackage{algorithmic}
\usepackage{graphicx}
\usepackage{textcomp}
\usepackage{xspace} 
\usepackage{todonotes}
\usepackage{subcaption}
\usepackage{multirow}
\usepackage{booktabs}
\usepackage{xcolor}
\usepackage{hyperref}

\def \alname{N2D\xspace}
\title{N2D: (Not Too) Deep Clustering via Clustering the Local Manifold of an Autoencoded Embedding}

\author{\IEEEauthorblockN{Ryan McConville, Ra\'ul Santos-Rodr\'iguez, Robert J Piechocki and Ian Craddock}
\IEEEauthorblockA{School of Computer Science, Electrical and Electronic Engineering, and Engineering Maths \\
University of Bristol,
United Kingdom\\
Email: \{ryan.mcconville, enrsr, r.j.piechocki, ian.craddock\}@bristol.ac.uk}
}

\begin{document}
\date{}

\maketitle

\begin{abstract}
Deep clustering has increasingly been demonstrating superiority over conventional shallow clustering algorithms. 
Deep clustering algorithms usually combine representation learning with deep neural networks to achieve this performance, typically optimizing a clustering and non-clustering loss.
In such cases, an autoencoder is typically connected with a clustering network, and the final clustering is jointly learned by both the autoencoder and clustering network.
Instead, we propose to learn an autoencoded embedding and then search this further for the underlying manifold.
For simplicity, we then cluster this with a shallow clustering algorithm, rather than a deeper network.
We study a number of local and global manifold learning methods on both the raw data and autoencoded embedding, concluding that UMAP in our framework is able to find the best clusterable manifold of the embedding. This suggests that local manifold learning on an autoencoded embedding is effective for discovering higher quality clusters.
We quantitatively show across a range of image and time-series datasets that our method has competitive performance against the latest deep clustering algorithms, including out-performing current state-of-the-art on several.
We postulate that these results show a promising research direction for deep clustering. The code can be found at \url{https://github.com/rymc/n2d}.

\end{abstract}

\section{Introduction}
Clustering is a fundamental pillar of unsupervised machine learning. 
It is widely used in a range of tasks across disciplines and well-known algorithms such as $k$-means have found success in many applications.
For example, in science, data exploration and understanding is a fundamental task which clustering facilitates by uncovering the hidden structure of the data.
However, $k$-means~\cite{Lloyd82leastsquares}, along with many conventional clustering algorithms such as Gaussian Mixture Models (GMMs)~\cite{Reynolds2009GaussianMM}, DBSCAN~\cite{Ester:1996:DAD:3001460.3001507}, and hierarchical algorithms~\cite{jain1999data} typically require hand engineered features to be created for each dataset and task. 
Further, these features may then be analysed using another process, feature selection, in order to eliminate redundant or poor quality features.
This task is even more challenging in the unsupervised setting.
Additionally, it is a time-consuming and brittle process, with the choice of features having a large influence over the subsequent performance of the clustering algorithm.

However, recent advances in deep learning have paved the way for algorithms which can effectively learn from raw data, bypassing the need for manual feature extraction and selection. 
One such popular method which learns powerful representations of the data automatically is an autoencoder~\cite{DBLP:journals/corr/abs-1812-05069}. Autoencoders effectively seek to learn the intrinsic structure of the data with a deep neural network, and do so by learning to reconstruct the original data, regularized for example, via a bottleneck inducing a compressed representation.
This representation learned from the raw data is then typically used in a range of tasks, such as an input to a supervised classifier.

This line of research has also impacted the unsupervised domain, where deep clustering has become a popular area of study. 
Deep clustering refers to the process of clustering with deep neural networks, typically with features automatically learned from the raw data by CNNs~\cite{yangCVPR2016joint} or autoencoders~\cite{pmlr-v48-xieb16} and clustered with a deep neural network.

These algorithms have reported large performance gains on various benchmark tasks over conventional non-deep clustering algorithms.
For example, Guo et al.~\cite{ijcai2017-243} pre-train an autoencoder, then initialize the weights of a deep clustering network with $k$-means.
Following this, the autoencoder continues to learn a representation, but jointly with the clustering network which seeks a good clustering.

In this work, we propose a simple approach, \alname, that effectively replaces the clustering network with a manifold learning technique on top of the autoencoded representation.  
Specifically, we intend for it to find a distance preserving manifold within this representation. 
Given this updated embedding, we can then cluster it with conventional non-deep clustering algorithms.
By doing so, \alname replaces the complexity of the clustering network with a manifold learning method and straightforward non-deep clustering algorithm, reducing the deepness of the deep clustering, yet achieving superior performance via the extra manifold learning step.

One important question is which manifold learning technique to apply to the autoencoded representation. There are many possible methods, such as the well-known Principal Component Analysis (PCA)~\cite{Abdi:2010:PCA:3160436.3160440}.
PCA seeks to learn a linear transformation of data into a new space, typically via the use of eigendecomposition of the covariance matrix, or by computing the Singular Value Decomposition (SVD) of the data.
However, PCA is a linear method and does not perform well in cases where relationships are non-linear.
Thankfully, alternative non-linear manifold learning methods exist, and can be categorised by their focus on finding local or global structure.
Well known globally focused methods include Isomap~\cite{tenenbaum2000global}, while t-SNE~\cite{maaten2008visualizing} is a well known locally focused method. 
More recently, UMAP~\cite{mcinnes2018umap} has been proposed, which while also local, has been shown to better preserve global structure. 
All of these methods seek to utilize the distances between points in order to better learn the underlying structure, and we posit that they will improve the clusterability of an autoencoded embedding.
To better understand this, we study the performance of each of these manifold learning methods on both the raw data and the autoencoded embedding. 

Thus, we propose a framework, \alname, where in contrast to recent deep clustering techniques, we replace the deep clustering network with a manifold learning method, and shallow cluster the resulting re-embedded space.
We empirically observe that this method is competitive (top-3) with state-of-the-art deep clustering algorithms across a range of datasets. Further, we observe that it out-performs state-of-the-art algorithms on several others. 
Code and weights to reproduce the results are available at \url{https://github.com/rymc/n2d}.

\section{Related Work}
Clustering algorithms can be broadly categorized into two different categories, hierarchical clustering and partitional clustering. 
Hierarchical clustering~\cite{johnson1967hierarchical} algorithms themselves can be categorized into divisive and agglomerative, where the former repeatedly splits clusters as it moves down the hierarchy, and the latter merges clusters as it moves up the hierarchy. 
{}artitional clustering algorithms are an alternative approach which divides a dataset into typically non-overlapping subsets at a single level.
Gaussian Mixture Models (GMMs)~\cite{Reynolds2009GaussianMM} and $k$-means~\cite{Lloyd82leastsquares} are two well-known partitional clustering algorithms. 
$k$-means divides a dataset into $k$ disjoint clusters by typically minimizing the sum of squared errors between each datapoint and their closest cluster centroid.
GMMs are a probabilistic model that can be considered as generalized $k$-means to utilize covariance structure information and the centers of the latent Gaussians.

As the performance of various machine learning algorithms, including clustering algorithms, is heavily dependent on the choice of features, much work has occurred in the area of automatically learning these features, or representations of the data. 
There exist deep learning based methods, such as autoencoders, which seek to autoencode a high dimensional mapping to a lower one, such that the higher dimensional mapping can be reconstructed again.
There is also significant work in non-deep methods such as PCA~\cite{Abdi:2010:PCA:3160436.3160440} , ICA~\cite{hyvarinen2000independent}, Local Linear Embedding (LLE)~\cite{roweis2000nonlinear} and Isomap~\cite{tenenbaum2000global}.
Methods such as PCA seek to preserve the important structure of the data, while other methods such as LLE and Isomap, which preserve the geometric and neighbour properties of the data.

A relatively recent area of study that combines both of these lines of research is deep clustering. Deep clustering methods use deep neural networks to cluster, typically involving two different processes, one where a representation is learned,  and one where the actual clustering occurs. 
This process may occur separately or jointly.

The deep neural networks used for deep clustering are diverse, and include MLPs~\cite{pmlr-v48-xieb16}, Convolutional Neural Networks (CNNs)~\cite{yangCVPR2016joint} and Generative Adversarial Networks (GANs)~\cite{Mukherjee2019ClusterGANL}. 
When used in the representation learning step these methods will optimize a specific loss, such as the reconstruction loss or generative adversarial loss. 
However, in addition, a clustering loss is added to guide the algorithm to find more cluster friendly features. These losses may include a $k$-means loss~\cite{yang2017towards} or a cluster hardening loss~\cite{pmlr-v48-xieb16}.
These losses are then typically combined in some way, such as with joint training, where the clustering loss is usually given much lower weight than the non-clustering loss~\cite{ijcai2017-243}.

 Along these lines, IDEC~\cite{ijcai2017-243} and ASPC-DA~\cite{8693526} both use an autoencoder for their initial pre-training step. Based on this learned representation, these methods initialize the weights of a new clustering network with $k$-means. IDEC and ASPC-DA then jointly trained this clustering network with the autoencoder. These approaches have been shown to perform well on a number of clustering tasks.

An alternative to using two different losses is to use a single combined loss, such as DEC~\cite{pmlr-v48-xieb16} or JULE~\cite{yangCVPR2016joint}. JULE uses CNNs as the representation learning step, integrating the learning of the representation and clustering into the backward and forward passes of a single recurrent model. The downside of their approach is that it is quite inefficient due to the recurrent nature of the model.

The concept of manifold learning on embeddings has been explored by Hasan and Curry~\cite{hasan-curry-2017-word}. In this work they specifically study the setting of applying LLE to existing word embeddings, improving the performance of word embeddings in word similarity tasks. 
They show how this method has theoretical foundations in metric recovery~\cite{TACL809}.
We note that in this work they apply LLE to windows of the embedding and use it to transform test vectors from the original embedding, whereas we are interested in learning the manifold of the entire embedding, optimizing for clusterability. 

Others have studied the integration of local constraints into autoencoder learning. 
Wei et al. ~\cite{10.1371/journal.pone.0146672} propose a semi-supervised method for document representation which utilizes an autoencoder to jointly learn from both the document itself, and neighbouring documents. They then demonstrate that by incorporating locality they improve performance in both document clustering and classification.

\section{Method}
Our method relies primarily on the combination of two different manifold learning methods.
The first is an autoencoder, which while learning a representation, does not explicitly take local structure into account.
We will show that by augmenting the autoencoder with a manifold learning technique which explicitly takes local structure into account, we can increase the quality of the representation learned in terms of clusterability.

\subsection{Autoencoder}
An autoencoder is a deep neural network consisting of two key components. 
The first is the encoder, which attempts to learn a function which maps the input $x$ to a new feature vector ($h = f(x)$).
The second component is the decoder, which attempts to learn a function which maps the learned feature space back to the original input space ($r = g(h)$. 
In other words, it is a neural network which attempts to copy its input to its output.
This is typically achieved via a form of regularization, for example by forcing the network to compress the input into a lower dimensional space.

The learning process can be described as minimizing the loss function $L(x, g(f(x)))$, where $L$ is a function which penalizes $g(f(x))$ for being dissimilar to $x$. One such loss may be the Mean Squared Error (MSE). 

While autoencoders have been shown to perform well at many feature representation tasks, they do not explicitly preserve the distances of the data in the representation that they learn. We believe that by taking this into account we can improve the quality of the clusters found.

\subsection{Isomap}
There are a multitude of manifold learning techniques that explicitly seek to preserve distances within the data. 
Isomap~\cite{tenenbaum2000global} is a nonlinear method which extends multidimensional scaling (MDS) to incorporate geodesic distances imposed by a weighted graph.  Geodesic distance is the distance between two points measured over the manifold, and thus by using the geodesic distance Isomap can learn the manifold structure. 
A k-nearest neighbourhood graph is constructed from the data, where the shortest distance between two nodes is considered the geodesic distance. 
Isomap constructs a global pairwise geodesic similarity matrix between all points in the data, on which classical scaling is applied.
Thus, Isomap can be considered a global manifold learning technique as it seeks to retain the global structure of the data. 
While Isomap is a global approach and our hypothesis is that a learning a local manifold on the autoencoded embedding will lead to better results, we will investigate the use of Isomap within \alname, specifically to understand how a global method performs and test our hypothesis.

We consider Isomap to have two key parameters in our setting, the first is the number of components, which is the top $n$ eigenvectors of the geodesic distance matrix which represent the co-ordinates in the new space.
The next parameter of importance is the number of neighbours to consider, which is simply the number of $k$-nearest neighbours to consider as local to a point.

\subsection{t-SNE}
t-SNE (t-distributed Stochastic Neighbor Embedding)~\cite{maaten2008visualizing} is a nonlinear method with a specific objective of optimizing local distances when creating the embedding.
The first stage of the t-SNE algorithm is to construct a probability distribution over pairs within the data in such away that similar points will have a high probability of being chosen while dissimilar points have an extremely low probability of being chosen.
In the second stage t-SNE defines a probability distribution over the mapped points, minimising the Kullback–Leibler (KL) divergence between the two distributions.

As with Isomap, t-SNE can choose the number of components in which to embed the data. 
It also requires a perplexity value which is related to the number of nearest neighbours used in Isomap. 
However, t-SNE is typically not very sensitive to this value.

\subsection{UMAP}
A recently proposed manifold learning method is UMAP (Uniform Manifold Approximation and Projection)~\cite{mcinnes2018umap}, which seeks to accurately represent local structure, but has been shown to also better incorporating global structure.
Compared to t-SNE it has a number of benefits to motivate the comparison with t-SNE in our framework.
While t-SNE typically struggles with large datasets, UMAP has been shown to scale well.
Further, as UMAP better preserves global structure, while remaining focused on preserving distances within local neighbourhoods, it may inherit benefits from both local and global methods.

UMAP relies on three assumptions, namely that the data is uniformly distributed on a Riemannian manifold, that the Riemannian metric is locally constant and that the manifold is locally connected.
From these assumptions it is possible to model the manifold with a fuzzy topological structure. The embedding is found by searching for a low dimensional projection of the data that has the closest possible equivalent fuzzy topological structure.

UMAP is similar to Isomap~\cite{tenenbaum2000global} in that it uses a k-neighbour based graph algorithm to compute the nearest neighbours of points.
At a high level, UMAP first constructs a weighted k-neighbour graph, and from this graph a low dimensional layout is computed.
This low dimensional layout is optimized to have as close a fuzzy topological representation to the original as possible based on cross entropy.

It has a number of important hyperparameters that influence performance. 
The first is the number of neighbours to consider as local.
This represents the trade-off between the granularity of how much local structure is preserved and how much of the global structure is captured.
As we are primarily concerned with the integration of local structure into our embedding, we will typically choose lower values for the number of neighbours.
The second is the dimensionality of the target embedding. 
In our method we set the dimensionality to be the number of clusters we are seeking to find.
UMAP also requires the minimum allowed separation between points in the embedding space.
Lower values of this minimum distance will more accurately capture the true manifold structure, but may lead to dense clouds that make visualization difficult.

\subsection{\alname}
We posit that by learning the manifold of the autoencoded embedding, specifically learning a manifold with a specific emphasis on locality, we can achieve a more cluster friendly embedding. 
However, as there is generally no ability to cross-validate hyperparameters in the unsupervised setting, it is therefore important to choose sensible default parameters for each approach.
For all manifold learning methods, we set the number of components or dimensions to be the number of clusters in the data.
For Isomap and UMAP we consider the number of neighbours to be an important parameter, and we set it to a sensible default value of 5 for Isomap, and 20 for UMAP.
UMAP also has another parameter we believe will be influential, which is the minimum distance between points. We believe that a default minimum distance of 0 is ideal for our method, as our prime motivation is not visualization and thus a more accurate representation of the true manifold is preferred.

We summarize the high level steps of our proposed method \alname as:
\begin{itemize}
    \item Apply an autoencoder to the raw data to learn an initial representation.
    \item We re-embed the autoencoded embedding by searching for a more clusterable manifold with a manifold learning method which preserves local distances.
    \item Finally, given this new, more clusterable embedding, we apply a final shallow clustering algorithm to discover the clusters.
\end{itemize}

More concisely, we may also simply represent \alname as 
\begin{equation}
    C = F_C(F_M(F_A(X)))
\end{equation}
where $C$ is the final clustering, $F_C$ is the clustering algorithm, $F_M$ is the manifold learner, $F_A$ is the autoencoder and $X$ is the original data.

We will study three manifold learning methods to understand the effect of the various approaches when applied to both the raw data and the autoencoded embedding, showing how one specific method, UMAP, achieves superior performance when applied to the embedding.
On the question of why combine an autoencoder with a manifold learning method, we will demonstrate empirically in Section \ref{sect:results} the contribution of each step to the overall performance, showing how this step can significantly increase performance.
We will also demonstrate in Section \ref{sect:results} how it is competitive with the state-of-the-art across a range of datasets, both image and time-series, and itself achieves state-of-the-art results on several.

\begin{figure*}

\begin{subfigure}[t]{0.33\textwidth}
    \centering
    \includegraphics[scale=.28]{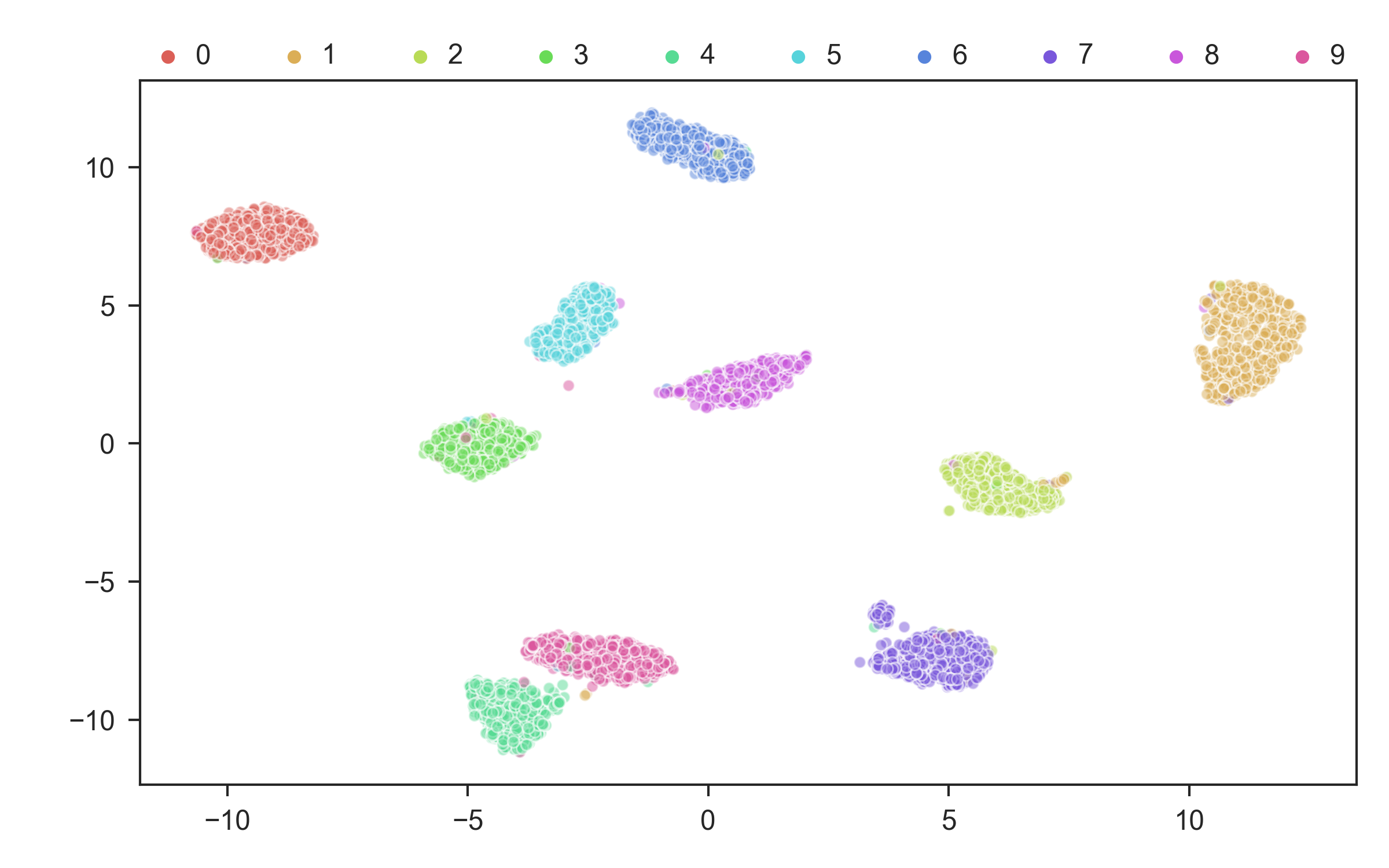}
    \caption{MNIST}
    \label{fig:mnist-n2d}
\end{subfigure}%
\begin{subfigure}[t]{0.33\textwidth}
    \centering
    \includegraphics[scale=.28]{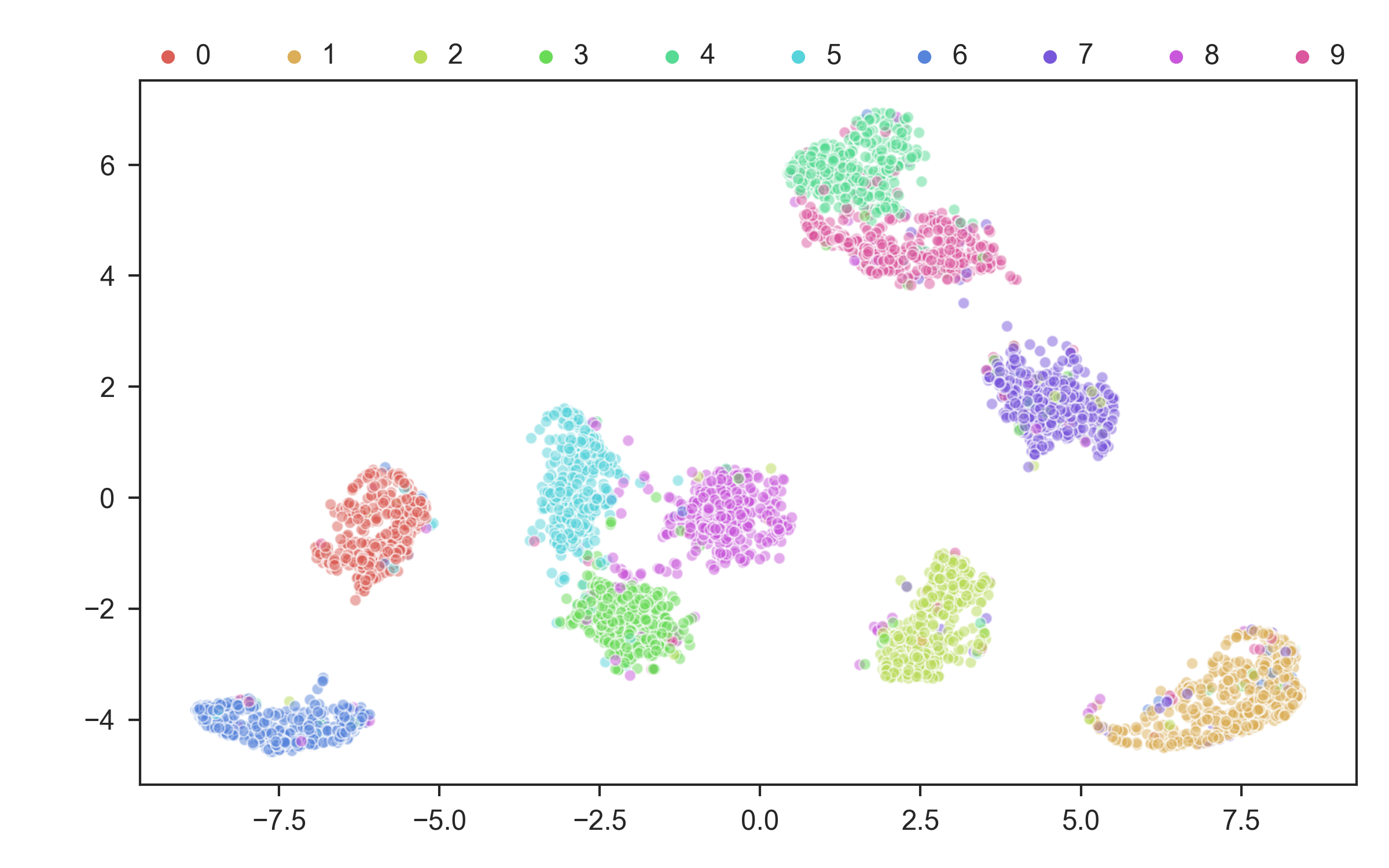}
    \caption{MNIST-test}
\end{subfigure}%
\begin{subfigure}[t]{0.33\textwidth}
    \centering
    \includegraphics[scale=.28]{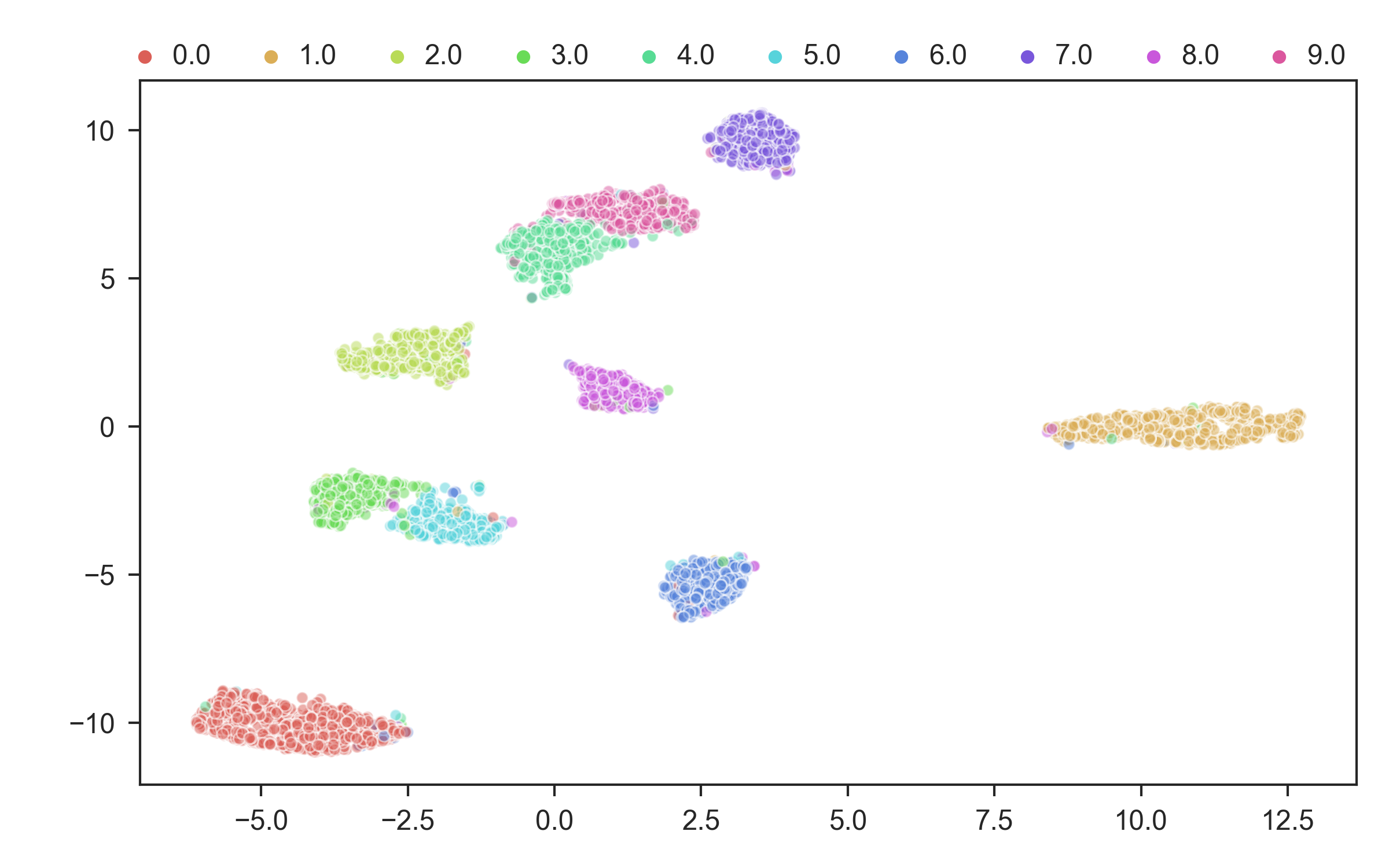}
    \caption{USPS}
    \label{fig:usps-n2d}
\end{subfigure}%

\begin{subfigure}[t]{0.33\textwidth}
    \centering
    \includegraphics[scale=.28]{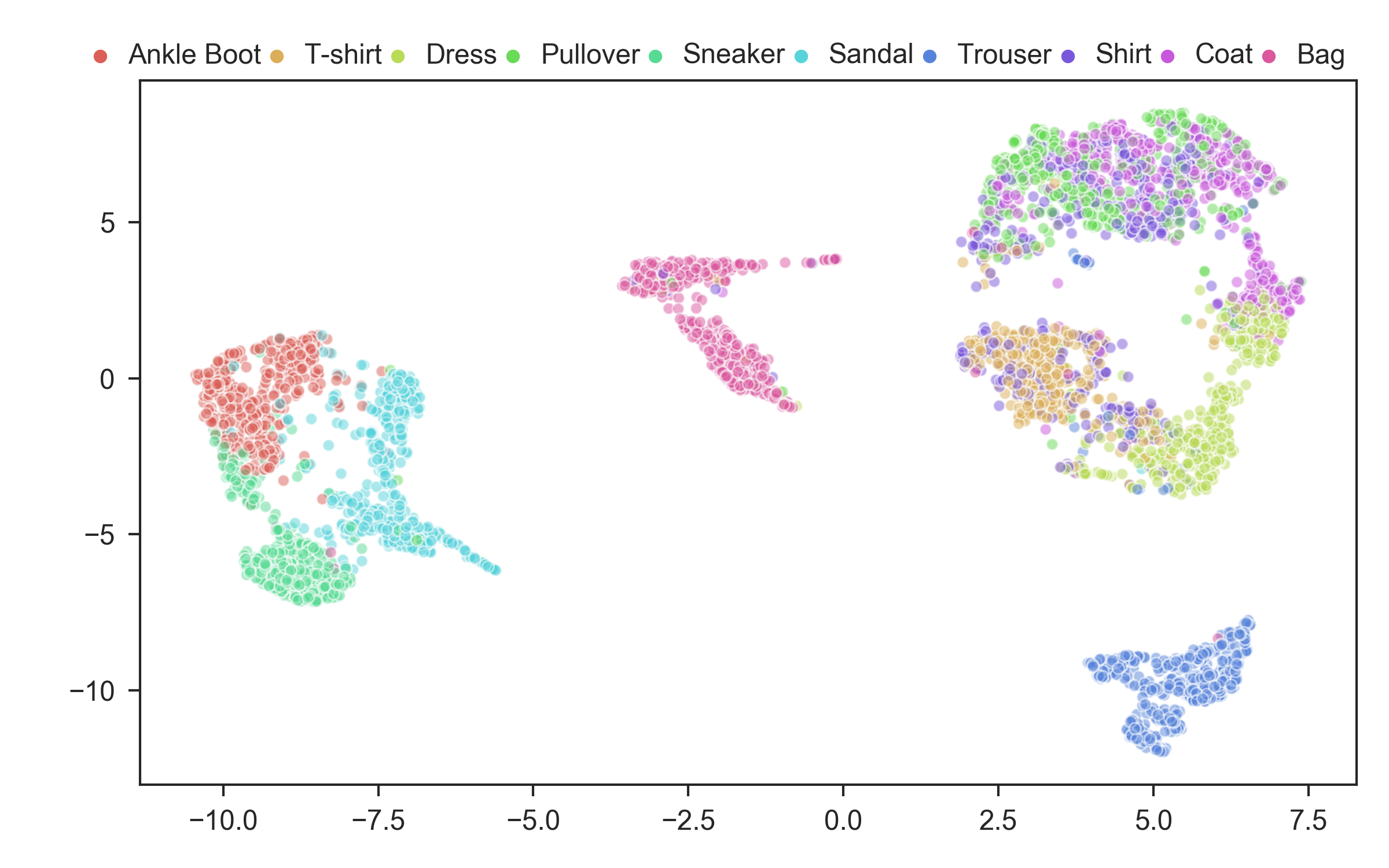}
    \caption{Fashion}
    \label{fig:fashion-n2d}
\end{subfigure}%
\begin{subfigure}[t]{0.33\textwidth}
    \centering
    \includegraphics[scale=.28]{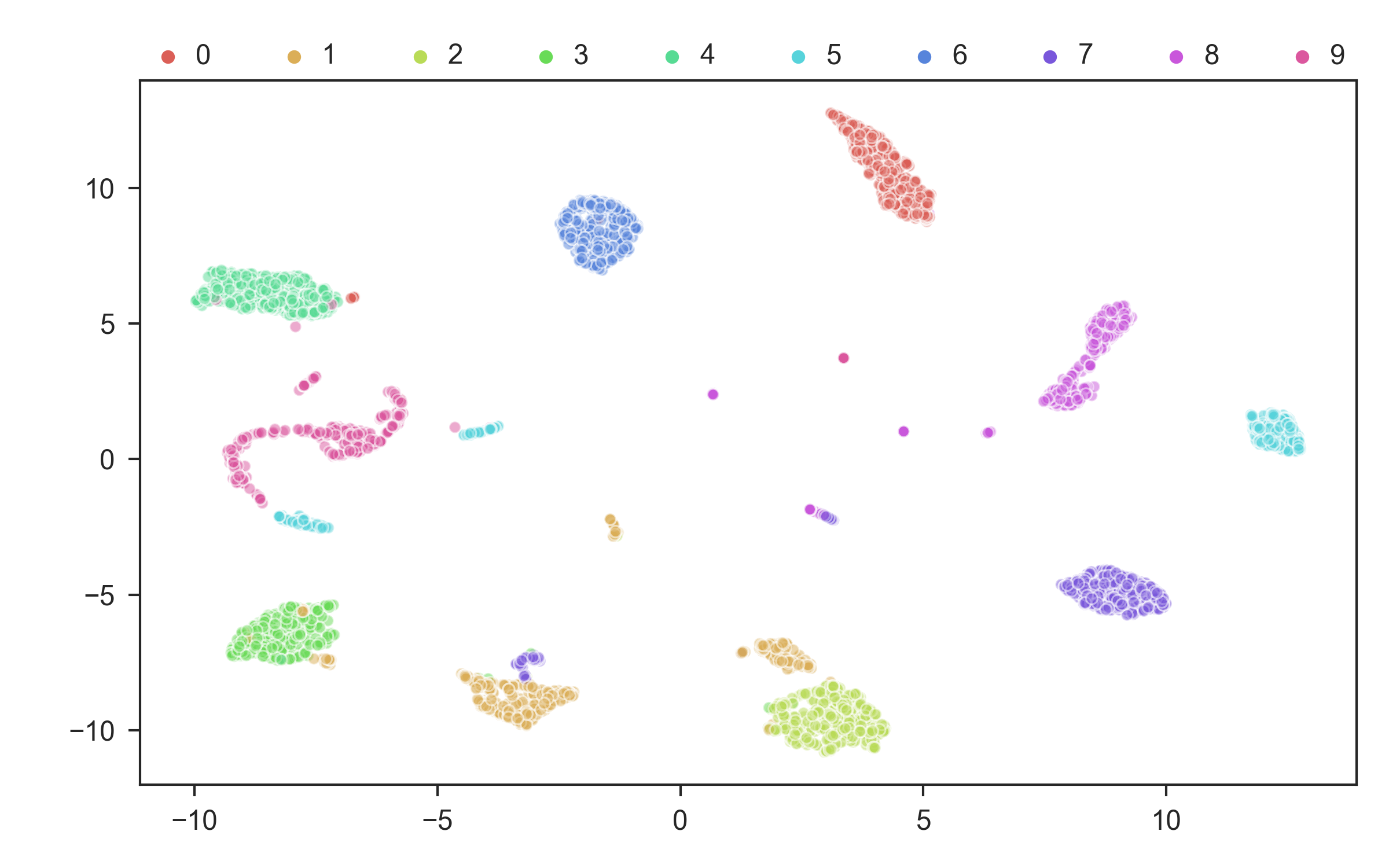}
    \caption{pendigits}
\end{subfigure}%
\begin{subfigure}[t]{0.33\textwidth}
    \centering
    \includegraphics[scale=.28]{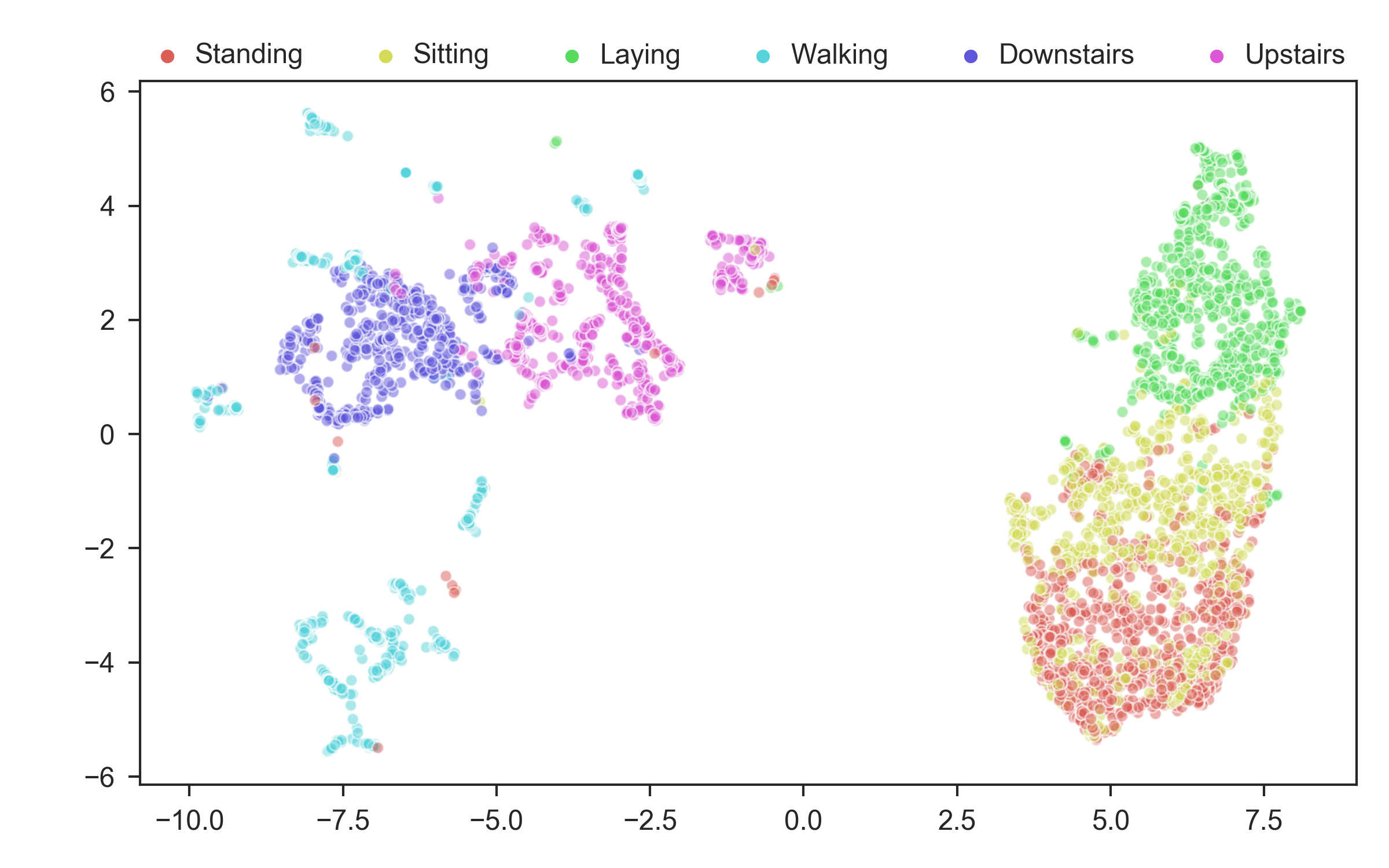}
    \caption{HAR}
            \label{fig-har-n2d}

\end{subfigure}
\caption{Visualization of \alname applied to all six datasets. For visualization purposes we set the number of dimensions to 2 and points plotted to 5000. In contrast, when we use \alname for clustering and not visualization, we cluster in higher dimensions (where the number of dimensions is the number of clusters) and achieve higher clustering performance.}
\label{lab:plots}
\end{figure*}

\section{Experiments}
In order to validate our idea, we conduct experiments on a range of diverse datasets, including standard datasets used to evaluate deep clustering algorithms.

\subsection{Datasets}
\begin{itemize}
    \item MNIST: A traditional benchmark dataset consisting of 70,000 handwritten digits belong to 10 different classes. 
    \item MNIST-test: A subset of the MNIST dataset, containing only the test set of 10,000 images.
    \item USPS: A dataset of 9298 images belonging to 10 different classes. Whereas MNIST images are 28x28, these images are 16x16.
    \item Fashion: A more challenging alternative to the MNIST dataset, consisting of 70,000 images of clothing, for a total of 10 classes.
    \item pendigits: A time series dataset consisting of sampled points from a pressure sensitive tablet as ten different digits are written. Each digit is represented by 8 coordinates of the stylus when writing a specific digit. There are 10992 data points.
    \item HAR: A time series dataset consisting of sensor data from a smart phone. It was collected from 30 people performing various activities of daily living, and contains 6 different activities; walking, walking upstairs, walking downstairs, sitting, standing and laying.

\end{itemize}

\begin{table*}[]
\centering
\caption{Evaluating the performance of each component of the proposed method.  AE, UMAP and \alname each have a GMM clustering step post-manifold learning. Results for Isomap were sometimes not available (\textemdash{}) as it exhausted memory on our 64GB machine.}

\resizebox{\textwidth}{!}{%
\begin{tabular}{|l|l|l|l|l|l|l|l|l|l|l|l|l|}
 & \multicolumn{2}{l|}{MNIST} & \multicolumn{2}{l|}{MNIST-test} & \multicolumn{2}{l|}{USPS} & \multicolumn{2}{l|}{Fashion} & \multicolumn{2}{l|}{pendigits} & \multicolumn{2}{l|}{HAR} \\
 & ACC & NMI & ACC & NMI & ACC & NMI & ACC & NMI & ACC & NMI & ACC & NMI \\
GMM & 0.389 & 0.333 & 0.464 & 0.465 & 0.562 & 0.541 & 0.463 & 0.514 & 0.674 & 0.683 & 0.585 & 0.648 \\
AE & 0.809 & 0.835 & 0.769 & 0.745 & 0.707 & 0.705 & 0.569 & 0.591 & 0.807 & 0.757 & 0.552 & 0.483 \\
Isomap & \textemdash{} & \textemdash{} & 0.896 & 0.755 & 0.781 & 0.787 & \textemdash{} & \textemdash{} & 0.756 & 0.792 & 0.608 & 0.677 \\
\alname (Isomap) & \textemdash{} & \textemdash{} & 0.797 & 0.77 & 0.660 & 0.727 & \textemdash{} & \textemdash{} & 0.826 & 0.820 & 0.632 & 0.580 \\
TSNE & 0.768 & 0.810 & 0.806 & 0.823 & 0.720 & 0.830 & 0.608  & 0.651  & 0.893 & 0.872 & 0.647 & 0.736 \\
\alname (TSNE) & 0.978 & 0.940 & 0.948 & 0.883 & 0.810 & 0.858 & 0.588  & 0.658 & 0.785 & 0.826 & 0.768 & 0.670 \\
UMAP & 0.825 & 0.880 & 0.857 & 0.819 & 0.804 & 0.845 & 0.588 & 0.656 & 0.819 & 0.856 & 0.552 & 0.696 \\
\alname (UMAP) & 0.979 & 0.942 & 0.948 & 0.882 & 0.958 & 0.901 & 0.672 & 0.684 & 0.885 & 0.863 & 0.801 & 0.683
\end{tabular}%
}
\label{tab:which_parts}
\end{table*}

\subsection{Evaluation Metrics}
We will use two standard evaluation metrics for validating the performance of unsupervised clustering algorithms.
In both cases, values range between 0 and 1, where higher values correspond to better clustering performance.

\subsubsection{Accuracy}
In clustering, accuracy (ACC) is defined as the best match between the ground truth and the predicted clusters. 
\begin{equation}
  ACC = \max_m \frac{\sum_{i=1}^n \mathbf{1}\{y_i = m(c_i)\}}{n}
\end{equation}
where $y$ are the ground truth labels, $c$ are the cluster labels, and $m$ enumerates mappings between clusters and labels.
 
\subsubsection{Normalized Mutual Information}
The Normalized Mutual Information (NMI) can be viewed as a normalization of the mutual information to scale the results between 0 and 1, where 0 has no mutual information and 1 is perfect correlation. More concretely, NMI is defined as:
\begin{equation}
NMI = \frac{
2I(y, c)}{[H(y)+ H(c)]}
\end{equation}
where $y$ are the ground truth labels, $c$ are the cluster labels, $H$ measures the entropy, and $I$ is the mutual information between the ground truth labels and the cluster labels.

\subsection{Experimental Settings}
We base our autoencoder on the architecture described by Xie et al.~\cite{pmlr-v48-xieb16}, which is a fully connected Multi-Layer Perceptron (MLP). 
The dimensions are inspired by those chosen by van der Maaten et al. in t-SNE~\cite{maaten2008visualizing}, which are $d$-500-500-2000-$c$, where $d$ is the dimensionality of the data and $c$ is the number of clusters.
As typical with autoencoders, the decoder network is a mirror of the encoder. All layers use ReLU activation~\cite{relu}. The optimizer is Adam~\cite{kingma2014adam}.
We train the autoencoder on for 1000 epochs for all datasets.

We use UMAP with the following default parameter set across all datasets. The number of neighbours is 20, the number of dimensions is the number of clusters, and the minimum distance between each point in the manifold is 0.

We use a GMM for the final clustering algorithm, where each component has its own general covariance matrix, and there are $c$ components, where $c$ is the number of clusters.
 
 \begin{table*}[ht]
\centering
\caption{A comparison of our method with both shallow clustering algorithms, along with the latest deep-clustering algorithms. Results were retrieved from the literature, or computed by us when not found and possible to compute. The top 3 performing scores are highlighted in bold. Note that our method is consistently in the top-3 across 5 of the 6 datasets, including the best accuracy and NMI for Fashion, pendigits and HAR. Algorithms that are missing scores for datasets are because the paper did not originally test on this dataset and it was not easily possible to get this score.}
\label{tab:sota}
\resizebox{\textwidth}{!}{%
\begin{tabular}{lllllllllllll}
 & \multicolumn{2}{l|}{MNIST} & \multicolumn{2}{l|}{MNIST-test} & \multicolumn{2}{l|}{USPS} & \multicolumn{2}{l|}{Fashion} & \multicolumn{2}{l|}{pendigits} & \multicolumn{2}{l|}{HAR} \\
\multicolumn{1}{|l|}{} & \multicolumn{1}{l|}{ACC} & \multicolumn{1}{l|}{NMI} & \multicolumn{1}{l|}{ACC} & \multicolumn{1}{l|}{NMI} & \multicolumn{1}{l|}{ACC} & \multicolumn{1}{l|}{NMI} & \multicolumn{1}{l|}{ACC} & \multicolumn{1}{l|}{NMI} & \multicolumn{1}{l|}{ACC} & \multicolumn{1}{l|}{NMI} & \multicolumn{1}{l|}{ACC} & \multicolumn{1}{l|}{NMI} \\
\multicolumn{1}{|l|}{$k$-means~\cite{Lloyd82leastsquares}} & \multicolumn{1}{l|}{0.532} & \multicolumn{1}{l|}{0.450} & \multicolumn{1}{l|}{0.546} & \multicolumn{1}{l|}{0.501} & \multicolumn{1}{l|}{0.668} & \multicolumn{1}{l|}{0.626} & \multicolumn{1}{l|}{0.474} & \multicolumn{1}{l|}{0.512} & \multicolumn{1}{l|}{0.666} & \multicolumn{1}{l|}{0.681} & \multicolumn{1}{l|}{\textbf{0.599}} & \multicolumn{1}{l|}{0.588} \\
\multicolumn{1}{|l|}{SC~\cite{ng2002spectral}} & \multicolumn{1}{l|}{0.680} & \multicolumn{1}{l|}{0.759} & \multicolumn{1}{l|}{0.667} & \multicolumn{1}{l|}{0.712} & \multicolumn{1}{l|}{0.656} & \multicolumn{1}{l|}{0.796} & \multicolumn{1}{l|}{0.551} & \multicolumn{1}{l|}{\textbf{0.630}} & \multicolumn{1}{l|}{0.724} & \multicolumn{1}{l|}{\textbf{0.784}} & \multicolumn{1}{l|}{0.538} & \multicolumn{1}{l|}{\textbf{0.741}} \\
\multicolumn{1}{|l|}{GMM~\cite{Reynolds2009GaussianMM}} & \multicolumn{1}{l|}{0.389} & \multicolumn{1}{l|}{0.333} & \multicolumn{1}{l|}{0.464} & \multicolumn{1}{l|}{0.465} & \multicolumn{1}{l|}{0.562} & \multicolumn{1}{l|}{0.540} & \multicolumn{1}{l|}{0.463} & \multicolumn{1}{l|}{0.514} & \multicolumn{1}{l|}{0.673} & \multicolumn{1}{l|}{0.682} & \multicolumn{1}{l|}{0.585} & \multicolumn{1}{l|}{\textbf{0.648}} \\
\multicolumn{1}{|l|}{DeepCluster~\cite{caron2018deep}} & \multicolumn{1}{l|}{0.797} & \multicolumn{1}{l|}{0.661} & \multicolumn{1}{l|}{0.854} & \multicolumn{1}{l|}{0.713} & \multicolumn{1}{l|}{0.562} & \multicolumn{1}{l|}{0.54} & \multicolumn{1}{l|}{0.542} & \multicolumn{1}{l|}{0.510} & \multicolumn{1}{l|}{\textemdash{}} & \multicolumn{1}{l|}{\textemdash{}} & \multicolumn{1}{l|}{\textemdash{}} & \multicolumn{1}{l|}{\textemdash{}} \\
\multicolumn{1}{|l|}{DCN~\cite{yang2017towards}} & \multicolumn{1}{l|}{0.830} & \multicolumn{1}{l|}{0.810} & \multicolumn{1}{l|}{0.802} & \multicolumn{1}{l|}{0.786} & \multicolumn{1}{l|}{0.688} & \multicolumn{1}{l|}{0.683} & \multicolumn{1}{l|}{0.501} & \multicolumn{1}{l|}{0.558} & \multicolumn{1}{l|}{0.72} & \multicolumn{1}{l|}{0.69} & \multicolumn{1}{l|}{\textemdash{}} & \multicolumn{1}{l|}{\textemdash{}} \\
\multicolumn{1}{|l|}{DEC~\cite{pmlr-v48-xieb16}} & \multicolumn{1}{l|}{0.863} & \multicolumn{1}{l|}{0.834} & \multicolumn{1}{l|}{0.856} & \multicolumn{1}{l|}{0.830} & \multicolumn{1}{l|}{0.762} & \multicolumn{1}{l|}{0.767} & \multicolumn{1}{l|}{0.518} & \multicolumn{1}{l|}{0.546} & \multicolumn{1}{l|}{0.701} & \multicolumn{1}{l|}{0.678} & \multicolumn{1}{l|}{0.565} & \multicolumn{1}{l|}{0.587} \\
\multicolumn{1}{|l|}{IDEC~\cite{ijcai2017-243}} & \multicolumn{1}{l|}{0.881} & \multicolumn{1}{l|}{0.867} & \multicolumn{1}{l|}{0.846} & \multicolumn{1}{l|}{0.802} & \multicolumn{1}{l|}{0.761} & \multicolumn{1}{l|}{0.785} & \multicolumn{1}{l|}{0.529} & \multicolumn{1}{l|}{0.557} & \multicolumn{1}{l|}{\textbf{0.784}} & \multicolumn{1}{l|}{0.723} & \multicolumn{1}{l|}{\textbf{0.642}} & \multicolumn{1}{l|}{0.609} \\
\multicolumn{1}{|l|}{SR-$k$-means~\cite{jabi2018deep}} & \multicolumn{1}{l|}{0.939} & \multicolumn{1}{l|}{0.866} & \multicolumn{1}{l|}{0.863} & \multicolumn{1}{l|}{0.873} & \multicolumn{1}{l|}{0.901} & \multicolumn{1}{l|}{\textbf{0.912}} & \multicolumn{1}{l|}{0.507} & \multicolumn{1}{l|}{0.548} & \multicolumn{1}{l|}{\textemdash{}} & \multicolumn{1}{l|}{\textemdash{}} & \multicolumn{1}{l|}{\textemdash{}} & \multicolumn{1}{l|}{\textemdash{}} \\
\multicolumn{1}{|l|}{VaDE~\cite{vade}} & \multicolumn{1}{l|}{0.945} & \multicolumn{1}{l|}{0.876} & \multicolumn{1}{l|}{0.287} & \multicolumn{1}{l|}{0.287} & \multicolumn{1}{l|}{0.566} & \multicolumn{1}{l|}{0.512} & \multicolumn{1}{l|}{0.578} & \multicolumn{1}{l|}{\textbf{0.630}} & \multicolumn{1}{l|}{\textemdash{}} & \multicolumn{1}{l|}{\textemdash{}} & \multicolumn{1}{l|}{\textemdash{}} & \multicolumn{1}{l|}{\textemdash{}} \\
\multicolumn{1}{|l|}{ClusterGAN~\cite{Mukherjee2019ClusterGANL}} & \multicolumn{1}{l|}{0.964} & \multicolumn{1}{l|}{0.921} & \multicolumn{1}{l|}{\textemdash{}} & \multicolumn{1}{l|}{\textemdash{}} & \multicolumn{1}{l|}{\textemdash{}} & \multicolumn{1}{l|}{\textemdash{}} & \multicolumn{1}{l|}{\textbf{0.630}} & \multicolumn{1}{l|}{\textbf{0.640}} & \multicolumn{1}{l|}{\textbf{0.770}} & \multicolumn{1}{l|}{\textbf{0.730}} & \multicolumn{1}{l|}{\textemdash{}} & \multicolumn{1}{l|}{\textemdash{}} \\
\multicolumn{1}{|l|}{JULE~\cite{yangCVPR2016joint}} & \multicolumn{1}{l|}{0.964} & \multicolumn{1}{l|}{0.913} & \multicolumn{1}{l|}{\textbf{0.961}} & \multicolumn{1}{l|}{\textbf{0.915}} & \multicolumn{1}{l|}{\textbf{0.950}} & \multicolumn{1}{l|}{\textbf{0.913}} & \multicolumn{1}{l|}{0.563} & \multicolumn{1}{l|}{0.608} & \multicolumn{1}{l|}{\textemdash{}} & \multicolumn{1}{l|}{\textemdash{}} & \multicolumn{1}{l|}{\textemdash{}} & \multicolumn{1}{l|}{\textemdash{}} \\
\multicolumn{1}{|l|}{DEPICT~\cite{ghasedi2017deep}} & \multicolumn{1}{l|}{0.965} & \multicolumn{1}{l|}{0.917} & \multicolumn{1}{l|}{\textbf{0.963}} & \multicolumn{1}{l|}{\textbf{0.915}} & \multicolumn{1}{l|}{0.899} & \multicolumn{1}{l|}{0.906} & \multicolumn{1}{l|}{0.392} & \multicolumn{1}{l|}{0.392} & \multicolumn{1}{l|}{\textemdash{}} & \multicolumn{1}{l|}{\textemdash{}} & \multicolumn{1}{l|}{\textemdash{}} & \multicolumn{1}{l|}{\textemdash{}} \\
\multicolumn{1}{|l|}{DBC~\cite{li2018discriminatively}} & \multicolumn{1}{l|}{0.964} & \multicolumn{1}{l|}{0.917} & \multicolumn{1}{l|}{\textemdash{}} & \multicolumn{1}{l|}{\textemdash{}} & \multicolumn{1}{l|}{\textemdash{}} & \multicolumn{1}{l|}{\textemdash{}} & \multicolumn{1}{l|}{\textemdash{}} & \multicolumn{1}{l|}{\textemdash{}} & \multicolumn{1}{l|}{\textemdash{}} & \multicolumn{1}{l|}{\textemdash{}} & \multicolumn{1}{l|}{\textemdash{}} & \multicolumn{1}{l|}{\textemdash{}} \\
\multicolumn{1}{|l|}{DAC~\cite{chang2017deep}} & \multicolumn{1}{l|}{\textbf{0.978}} & \multicolumn{1}{l|}{\textbf{0.935}} & \multicolumn{1}{l|}{\textemdash{}} & \multicolumn{1}{l|}{\textemdash{}} & \multicolumn{1}{l|}{\textemdash{}} & \multicolumn{1}{l|}{\textemdash{}} & \multicolumn{1}{l|}{\textemdash{}} & \multicolumn{1}{l|}{\textemdash{}} & \multicolumn{1}{l|}{\textemdash{}} & \multicolumn{1}{l|}{\textemdash{}} & \multicolumn{1}{l|}{\textemdash{}} & \multicolumn{1}{l|}{\textemdash{}} \\
\multicolumn{1}{|l|}{ASPC-DA~\cite{8693526}} & \multicolumn{1}{l|}{\textbf{0.988}} & \multicolumn{1}{l|}{\textbf{0.966}} & \multicolumn{1}{l|}{\textbf{0.973}} & \multicolumn{1}{l|}{\textbf{0.936}} & \multicolumn{1}{l|}{\textbf{0.982}} & \multicolumn{1}{l|}{\textbf{0.951}} & \multicolumn{1}{l|}{\textbf{0.591}} & \multicolumn{1}{l|}{\textbf{0.654}} & \multicolumn{1}{l|}{\textemdash{}} & \multicolumn{1}{l|}{\textemdash{}} & \multicolumn{1}{l|}{\textemdash{}} & \multicolumn{1}{l|}{\textemdash{}} \\
\multicolumn{1}{|l|}{\alname} & \multicolumn{1}{l|}{\textbf{0.979}} & \multicolumn{1}{l|}{\textbf{0.942}} & \multicolumn{1}{l|}{0.948} & \multicolumn{1}{l|}{0.882} & \multicolumn{1}{l|}{\textbf{0.958}} & \multicolumn{1}{l|}{0.901} & \multicolumn{1}{l|}{\textbf{0.672}} & \multicolumn{1}{l|}{\textbf{0.684}} & \multicolumn{1}{l|}{\textbf{0.885}} & \multicolumn{1}{l|}{\textbf{0.863}} & \multicolumn{1}{l|}{\textbf{0.801}} & \multicolumn{1}{l|}{\textbf{0.683}}
\end{tabular}%
l}
\end{table*}
\begin{table}[]
\centering
\caption{Showing the efficiency of each stage of the method on each dataset, in minutes. AE refers to the autoencoder architecture defined earlier, trained for 1000 epochs on a Nvidia RTX 2080 Ti, and Manifold refers to manifold learning with UMAP. We do not use early stopping with the autoencoder training.}
\begin{tabular}{|l|l|l|l|}
 & AE (m) & Manifold (m) & Total (m) \\
MNIST & 18.0 & 1.5 & 19.5 \\
MNIST-test & 2.6 & 0.4 & 3.0 \\
USPS & 2.1 & 0.4  & 2.5 \\
Fashion &  18.0 & 1.5  & 19.5  \\
pendigits & 2.2 & 0.3 & 2.5 \\
HAR & 3.6 & 0.2 & 3.8
\end{tabular}
\label{tab:timings}
\end{table}

\subsection{Results}\label{sect:results}
Figure \ref{lab:plots} shows the resulting clusters when using \alname for visualization purposes. 
However, in order to better understand the effectiveness of our method at clustering we will study each individual component of \alname via measuring the accuracy and NMI, as well as how the full \alname algorithm compares with a range of other clustering algorithms.
\subsection{Role of Each Component of \alname}
Table \ref{tab:which_parts} shows details of the accuracy and NMI of each individual component of \alname.
This table shows that the performance of the non-deep clustering algorithm GMM is typically poorest across all datasets.
When we introduce manifold learning methods, and cluster those embeddings, we see improvements in cluster accuracy and NMI. 

As well as the autoencoder, we use 3 different manifold learning methods with different properties.
The first is Isomap, which is a globally focused manifold learner. 
It outperforms t-SNE, the local manifold learning technique, on 2 of the 4 datasets it was able to process. 
On 2 of the 6 datasets it was unable to complete the learning as it exhausted all memory on our 64GB system.
Therefore, on two datasets t-SNE performed better than Isomap, and on two others, Isomap outperformed t-SNE.

However, when Isomap and t-SNE are each applied to the autoencoded embedding, on only 1 of the 4 datasets does \alname with Isomap outperform \alname with t-SNE.
This suggests that on some datasets the clusters are better discovered by a global method, and others by a local method.
However, when applied the autoencoded embedding, the more local methods appear to be the better choice.

This intuitively suggests that a technique which is primarily locally focused but captures global structure better than t-SNE may lead to further improvements.
Therefore, when we experiment with UMAP, which meets this criteria, we see that UMAP is the superior approach on 3 of the 6 raw datasets.
However, when applied to the autoencoded embedding, \alname with UMAP outperforms both Isomap and t-SNE on all datasets.
This supports the hypothesis that a manifold learner, which, while locally focused, also captures a degree of the global structure, is best suited for discovering the clusterable manifold of an autoencoded embedding.

The largest gains between our approach \alname and the sub-components is on HAR, where there is a 25 percentage point increase in performance compared to the AE and UMAP, while on MNIST and USPS, where there is an around a 15 percentage point increase in accuracy when using \alname.

In Table \ref{tab:timings} we show the amount of time it takes for each stage of the method in minutes, as well as the total time. 
From this, it is clear that our method is efficient, clustering MNIST and Fashion-MNIST in around 18 minutes, while clustering the remaining 4 datasets in between two and four minutes.

\subsection{Comparison with other methods}
In Table \ref{tab:sota} we show the accuracy and NMI results for a wide set of clustering algorithms on six different datasets. 
The clustering algorithms chosen include a number of conventional non-deep methods, such as $k$-means, spectral clustering (SC) and GMMs.
They also include recent deep-clustering based methods, such as ClusterGAN, IDEC, JULE and ASPC-DA. 
These methods make significant use of deep networks, and typically outperform the non-deep clustering methods.

The most similar methods to \alname are IDEC and ASPC-DA. Both of these approaches pre-train an autoencoder before jointly training a second deep network with a clustering and non-clustering (reconstruction) loss. The clustering network weights are initialized with a non-deep clustering algorithm such as $k$-means.

In contrast, we replace the second deep network with a manifold learning method, UMAP, and then use a non-deep clustering algorithm, a GMM, to cluster the resulting embedding. 
Hence, our less deep method, \alname, benefits from less complexity, but as can be seen in Table \ref{tab:sota}, has competitive or superior performance to all other methods.

On five of the six datasets tested, our approach is in the top 3 for at least one of the metrics. 
On MNIST-test we are around 1 percentage point lower in accuracy than JULE and DEPICT, and 2 percentage points lower than ASPC-DA which is top.
However, on the Fashion dataset, we achieve the highest accuracy, around 5 absolute percentage points higher than ClusterGAN, and 8 absolute percentage points higher than ASPC-DA.

We also include two non-image datasets, pendigits and HAR, to validate performance on different types of data. Many of the best-performing deep-clustering methods are intended for image clustering (e.g., JULE~\cite{yangCVPR2016joint}, DBC~\cite{li2018discriminatively}, DAC~\cite{chang2017deep}), and thus we were unable to find or easily obtain results on these datasets. 
However, for the algorithms for which we could obtain or produce results, our method also achieved the best performance. 
For both datasets we compare our method with some of the most similar deep clustering approaches, DEC and IDEC. 
On pendigits, we achieve 11 percentage points higher accuracy than the closest approach IDEC and on HAR a 15 percentage point increase in accuracy.
In fact, consistently across all datasets, we achieve higher accuracy and NMI scores than these methods.

We also note that one of the closest competitors, ASPC-DA, which typically slightly outperforms our method on several datasets, achieves this performance due to data augmentation. 
When data augmentation is removed from ASPC-DA, they typically achieve less competitive performances, e.g. an accuracy of 0.924 (vs 0.988) on MNIST, 0.785 (vs 0.973) on MNIST-test and 0.688 (vs 0.982) on USPS. 
For future work we would like to evaluate our proposed method with data augmentation.

\section{Conclusion}
In this paper we propose a simple deep clustering method,  \alname, which reduces the deepness of typical deep clustering algorithms by replacing the clustering network with an alternative framework which seeks to find the manifold within the autoencoder embedding, and clusters this new embedding with a shallow clustering architecture.
We studied both global and local manifold learning algorithms, with our results supporting the hypothesis that learning the local manifold of an autoencoded embedding, while also preserving global structure as UMAP does, is better able to discover the most clusterable manifold of an autoencoded embedding.
\alname is the resulting combination which is shown to be effective on a range of datasets, including image and time-series datasets.
We compare \alname with both conventional shallow clustering algorithms, and the latest state-of-the-art deep clustering algorithms. 
In the empirical comparison, we show how our proposed method is competitive with the current state-of-the-art clustering approaches, achieving top-3 performance in five of the six datasets datasets tested.
Further, we outperform the state-of-the-art on several datasets, including surpassing the next best algorithm by around 5 absolute percentage points in accuracy on Fashion-MNIST and 15 percentage points on the activity recognition dataset HAR.

\bibliographystyle{IEEEtranS}
\bibliography{IEEEabrv,bib}
\end{document}